# Linguistic Classification using Instance-Based Learning

Priya S Nayak
*Dept. of Computer Science and Engineering*
*PES University*
Bengaluru, India
nayakpriya98@gmail.com

Rhythm Girdhar
*Dept. of Computer Science and Engineering*
*PES University*
Bengaluru, India
ridhigirdhar3@gmail.com

Shreekanth M Prabhu
*Dept. of Information Science and Engineering*
*CMR Institute of Technology*
Bengaluru, India
shreekanthpm@gmail.com

*Abstract*— Traditionally linguists have organized languages of the world as language families modelled as trees. In this work we take a contrarian approach and question the tree-based model that is rather restrictive. For example, the affinity that Sanskrit independently has with languages across Indo-European languages is better illustrated using a network model. We can say the same about inter-relationship between languages in India, where the inter-relationships are better discovered than assumed. To enable such a discovery, in this paper we have made use of instance-based learning techniques to assign language labels to words. We vocalize each word and then classify it by making use of our custom linguistic distance metric of the word relative to training sets containing language labels. We construct the training sets by making use of word clusters and assigning a language and category label to that cluster. Further, we make use of clustering coefficients as a quality metric for our research. We believe our work has the potential to usher in a new era in linguistics. We have limited this work for important languages in India. This work can be further strengthened by applying Adaboost for classification coupled with structural equivalence concepts of social network analysis.

## I. Introduction

European Visitors, in their 16th Century visit to India, noticed the commonality between Indian and European languages. This included English Jesuit Missionary Thomas Stephens [1], who wrote a letter in 1583 about similarities between Indian languages and Greek and Latin. Filippo Sassetti [2], writing in 1585, noted some word similarities between Sanskrit and Italian. However, neither of their observations led to further scholarly inquiry. There were others such as Marcus Zuerius van Boxhorn who noticed similarities among European and Persian languages and theorized common origin for them.

The topic resurfaced only in 1785, when Sir William Jones [3] gave his now memorable address to Asiatic Society. He recognized the close affinity between Sanskrit and European Languages and Persian Language which in turn led to the speculation of common origin for all the Indo-European Languages. His bold assertion contributed to the emergence of comparative linguistics and Indo-European Studies.

Many scholars were, initially, of the opinion that it was India which was the original home of Indo-European languages. Within a few decades however, the majority of scholars converged to the hypothesis that the common home was somewhere near Russian steppes. There was also a counter-view that the common home was Anatolia. This led to the hypothesis of Aryan Invasion theory. Later discovery of architectural finds near Mohenjo Daro and Harappa were explained by saying Harappans differed from Indo-Aryans and they were the target of Aryan Invasion. Other scholars, however, refuted this. There were no archeological findings to determine such an invasion. Overall, the controversy raged for the last two centuries. In recent times Michael Witzel [4] along with other linguists have claimed that Indo-Aryans were not autochthonous. Supporting the opposing view that Indians were autochthonous and maybe it was Indians who Aryanized Europe, were Shrikant Talageri [5], B. B. Lal [6] and Nicholas Kazanas [7]. Ed Bryant and Laurie Patton have compiled views on both sides about the Indo-Aryan controversy comprehensively [8].

In a paper presented by Priyadarshini [9] in 2010 made use of arguments spanning linguistics, archaeology and genetics that Indians are autochthonous to India. This controversy is yet to die. If the migrations happened into India, then it upholds the view of linguistics that Indian Languages can be boxed under Indo-Iranian branch of Indo-European language family. If we accept the opposing view, then we need to be open to a more network-like structure to represent inter-relationships.

The commonality between Indo-European speakers was not limited to language. It included traditions, myths and concepts of divinity and manifestations. The Vedic Gods namely Indra, Agni, Varuna, Mitra, Aryaman, Ashwins, Ushes, Dyaus and many others had counter-parts in European traditions. The work [10] details some of these parallels and similarities.

### A. Problem Definition

Originally, linguists depicted inter-relations between Indo-European languages in a tree structure. Jack Lynch [11] has depicted the languages as shown in Figure 1.

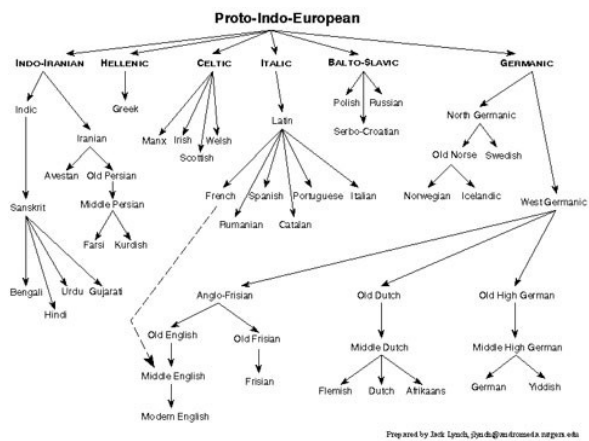

Fig. 1  Indo-European Language represented using Tree Model (Credits: Jack Lynch)

Based on the hierarchical structure, it shows that North Indian languages have similarities between them and many European languages. They are termed as Aryan languages whereas the South Indian languages show no connections with the European languages and are called Dravidian languages.

The above tree-based approach which involves a certain root and divergence of languages from that root was introduced by German Philologist August Schleicher [12]. This has remained main-stay for the comparative linguistics field to-date. However, when the languages are in contact and diversify, the wave model is proposed by researchers [13, 14].

In any given language there are words that are native to it or loaned from other languages. Indo-Aryan languages have words that are common to European languages as well as South Indian Languages or Dravidian Languages. Sanskrit has common words across Indian and European Languages. When a given word appears in an Indian language but not in European Language, some Linguists have considered it to be Dravidian. For example, the word - Mayura which refers to Peacock that is native to India is considered a Dravidian word. Such associations are questioned by those who are well-versed with literary sources in ancient India in particular Rigveda [15].

In general, it is hard to determine whether a given word is native to a language or a loan word. In most cases Sanskrit has retained the word root along with words derived from the root, whereas European languages have managed to retain only some of the derived words. Despite this, linguists have failed to give due to Sanskrit and instead gone about the search for an elusive Proto-Indo- European (PIE) Language which continues to remain in the realm of speculation.

Analysis even among Indian languages has to go beyond the simplistic Aryan and Dravidian divide. The word Ghee is *tuppa/toop* in Kannada, Marathi and Konkani whereas it is *nei* in Tamil, Malayalam and Telugu. In such cases it becomes harder to determine the source language and the source language family. According to Wiktionary [16] Tuppa is of Maharashtri Prakrat Origin. In fact, it was the Prakrats that gave rise to multiple Indian languages that are widely spoken today. This can be considered as an interim stage of language development between Sanskrit and today's languages.

In addition to Aryan and Dravidian languages there are a class of languages loosely termed as Munda languages. These languages are associated with Austro-Asiatic tribes in India. Linguists based on their understanding of how a typical word gets constructed in Sanskrit have considered words like Ganga as Munda words. They also consider the word for cotton - *kapas* as originally Munda word and Sanskriatized to *karpasam*. However, Indians are generally very familiar with Sanskrit words getting distorted in multiple languages with a phenomenon called *Apabhramsha*. Generally, Sanskrit is replete with words that require refined pronunciation and diction whereas colloquial languages use a simplified version of the same. Thus, the word *Shravan* is *Savan* in Hindi but retained as *Shravan* in many other languages. [17]

The study of languages is also important from the view-point of deciding the relative antiquity of a language. The aspect of antiquity has also been studied from the view-point of astronomy [18] and genetics [19, 20]. Literature suggests that there are changes in languages spreading as waves across geographies over time either due to physical migration or cultural transmission.

In summary, the objective of this work are as follows:

i.  Develop an approach to construct training sets that associate word(s) with a language using techniques from machine learning in place of techniques which linguists use that draw on linguistics domain. Since vocabulary of any language is typically gigantic, we focus on related word clusters.

ii. Classify words and assign language labels based on training sets. Here we consider local models.

iii. Reiterate between I and II till we get a model that has acceptable performance.

iv. Arrive at a new model for inter-relationship between languages that is more accurate than the current tree-based model.

The approach we intend to use is to model languages as clusters of word clusters and using the inter-relationships between word clusters across languages we enable discovery of inter-relationships between languages. We have restricted the scope to seven Indian languages, namely Sanskrit, Hindi, Punjabi, Marathi, Tamil, Telugu, Kannada, and English. One of the challenges in classifying words into languages is the prevalence of common words across Indian languages. Thus, we need approaches which are more refined than usual techniques.

Rest of the paper is as follows. In Section 2, we cover the *Related Work* where we dwell on techniques used by the Linguists to classify words in terms of the language they belong to and approach used by machine learning techniques. In Section 3, we cover the *Methodology* we use that combines unsupervised learning with instance-based learning in an iterative fashion. In Section 4, we present *Results*. Section 5 pertains to *Discussions* where we discuss the outcome of our exploration. Section 6, *Conclusions* concludes the paper and draws attention to opportunity for further experimentation.

## II. RELATED WORK

Bulk of research in comparative linguistics is native to the field. Linguists typically make use of a data-set of cognate words and their understanding of how languages get formed and evolve to make determination of the source language for a given word. However, linguists have to contend with other linguists who infer differently, as well as experts from other fields ranging from archeology to astronomy. In addition to cognate words, linguists make use of laws of parallelization and dialect distribution to draw inferences.

The edited volume on *Indo Aryan Controversy* [8] has arguments and counter-arguments by Satya Swarup Mishra [21], Koenraad Elst [22], Hans Henrich Hock [23], Subhash Kak [24], Shrikant G. Talageri [25], and Michael Witzel [26] look at the linguistic approach at understanding the Indo Aryan Controversy. However, this field can leverage machine learning approaches that can incorporate, automation and scale along with learning, which we concentrate on in this paper.

In this section we focus on application of machine learning to linguistics.

In recent times, phylogeny has been inter-linked with machine learning models in order to deduce inter relationships between languages. Phylogeny is the history of the progression of a species or group, especially regarding descent and relationships among vast groups of animals Phylogeography is the analysis of the principles and processes ruling the geographic distributions of developmental origins of organisms, within and among closely related species. [27] A Bayesian phylogeographic inference [28] treats language location as a continuous vector and deducts ancestral locations at internal nodes of the tree. Polish anthropologist Jan Czekanowski used a method of differential diagnosis by quantitative correlation determinations. [29] Using this as a foundation, Kroeber and Chrétien [30] attempt at classifying Indo-European languages. In their work, they speak about how the authenticity of the data, the number of groups involved creates clear distinctions in the results. Their paper looks at interrelationships between languages and moves on from the old tree-like structure that we have been familiar with until now.

Alix Boc et al., [31] proposed a similar approach. The assumption by Gray and Atkinson [32] that the evolution of languages was divergent and had a low frequency of borrowing of words was proven untrue. Only a network like structure, like the one mentioned above, can incorporate the borrowing and homoplasy (i.e., evolutionary convergence) processes. There was a proof that around 37% of the Indo-European words were affected by borrowing from other Indo-European Languages. They proposed to use a horizontal gene transfer detection method [33] in this bio-linguistics context to infer a phylogenetic network of the Indo-European Language family.

In the tree-structure shown in Fig. 2(a) and 2(b), one branch containing both Indian and Iranian languages showed the shortcomings of the entire structure and that led to multiple theories establishing that a network like structure is the most optimal way of representing it.

Another approach of classifying Indo-European language has been the introduction of Social Network Analysis [34]. Here they go with the hypothesis that if we build a network of languages (represented as word clusters) and establish linkages across languages based on word linkages the language which scores high on centrality measure probably has a greater chance of being the source language, older language or ancestral language.

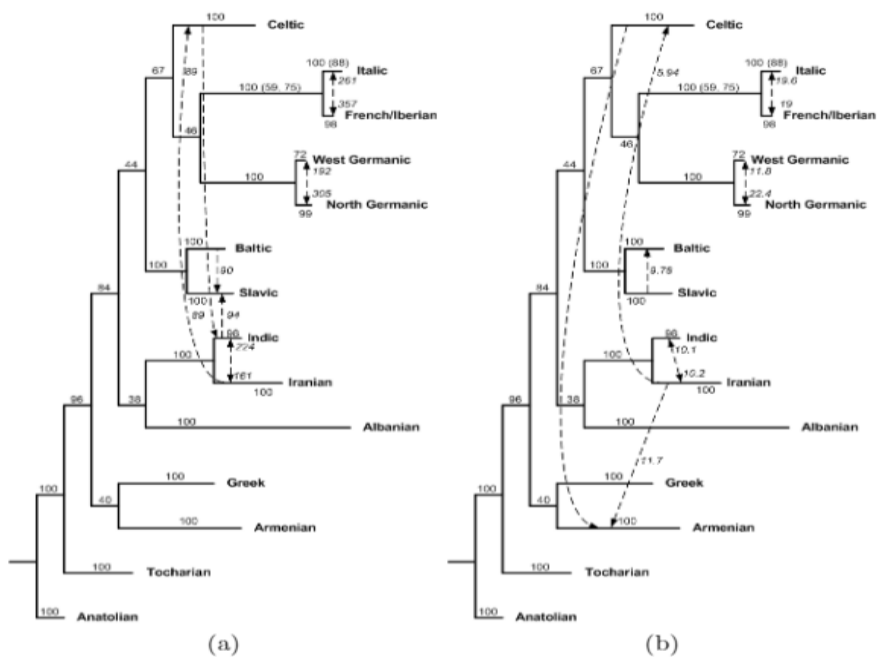

Fig. 2  Ten most frequent word exchanges between the IE language groups in terms of (a) total numbers of transferred words, and (b) percentages of affected words by group. (Credits: Alix Boc et. al)

In their paper, [35] Raghuveer Ketireddy and Kavi Narayana Murthy have presented their effort on automatic text classification in native Indian languages. Their paper uses Bayesian Learning Methods, K-Nearest Neighbor (KNN) Classifier. Soft-margin linear SVMs were used to test the performance. This analysis was one of the early papers that described the usage of machine learning in obtaining the inter-relationships between Indian Languages. They reported better results using SVM than other learning techniques. This approach has great computational complexity because of high dimensionality of the text vectors. Many researchers have used KNN for linguistic classification. However, KNN is prone to issues relating to high dimensionality.

Zhou et al. [36] research has paved the way for improving supervised learning algorithms by adding unsupervised elements. It suggests using K-means clustering along with traditional KNN. The proposed algorithm can lower the complexity of calculation and also improve the accuracy. However, the algorithm has some limitations w.r.t fixing the threshold value and determining the K value when clustering for each category. This work by Zhou et al., has validated that more research is needed that makes use of clustering as well as classification models.

Our work in this paper takes the work done by Zhou et al. forward by incorporating more improvisations and innovations for Indian languages.

### III. METHODOLOGY

We have used a methodical data collection approach and classification with clustering algorithms for the model. This is a more holistic approach by carefully selecting the best of all. Most researchers use transliteration directly but we have implemented vocalization like soundex to obtain more accurate results. We have used a custom distance measure which involves both Levenshtein and Euclidean distances. We have chosen to make use of the Density-based Spatial Clustering of Applications with Noise (DBSCAN) clustering algorithm instead of more widely-used k-means because it is robust and works with noisy data. The overview of the methodology is illustrated in Figure 3.

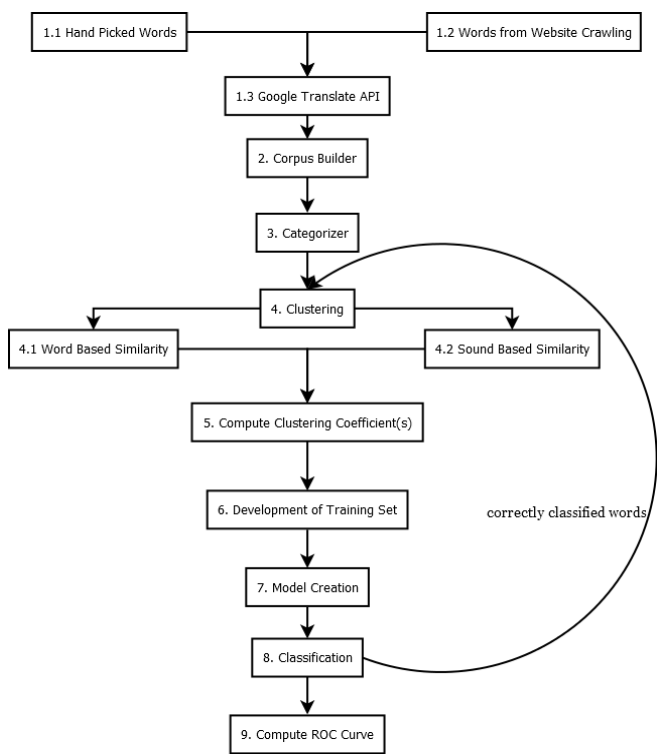

Fig. 3    Proposed Methodology for Linguistic Classification

Steps 1.1–3 concentrate on dataset creation, followed by Steps 4–5 involving dataset clustering to understand the unfamiliar words and categories that exist. Step 6 creates the training set using combining different clustering techniques. Step 7-8 works on the model based on custom metrics and KNN that classifies the words into a particular language. Correctly classified words are sent into the clustering module to improve the accuracy. Step 9 computes the Receiver Operating Curve (ROC) for the different classes once the classification module is frozen.

The following algorithms detail the steps that we have taken for each of the major modules. Algorithm 1 details the steps taken in building the dataset. Algorithm 2 talks about how words were clustered and the computation of the clustering coefficient(s). Algorithm 3 speaks about the approach taken in classifying a particular word.

**Algorithm 1: Creation of Dataset**

**procedure** CreateDataset ()

   **Output:** dataset consisting of different words, meanings, and soundex score created

   i. Choose handpicked words that are common to multiple languages.
   ii. Crawl websites to create a dataset that is in English
   iii. Use Google Translate to translate words into different languages
   iv. Use PyDictionary to derive meanings of different words
   v. Create categories by understanding the different words that are present in the dataset
   vi. Use Soundex by Python to retrieve the soundex score for a particular word

Steps i, ii, iii, and iv show the data creation involving handpicking words, website scraping, and then translating them to different languages, adding meanings to each word. Despite manual reviewing, it turned out that the same word (different pronunciation) across different languages had different meanings. This gave birth to the categorizer. The categorizer was done using human intelligence (Step v).

The English words were manually categorized and, like the meanings, were cascaded to the different languages. Additionally, each word was given a unique ID to distinguish it from the rest. Annexure 1 shows the dataset once the meanings and categories were added.

For Step vi, Phonetic score of a particular word can either be calculated using Soundex or NYSIIS. It was noted that NYSIIS performed well only on English words and didn't do justice to the rest of the corpus. Soundex distance is used to measure the phonetic distance between 2 words based on how they are spoken. This distance is measured by first converting the words into 4-character long soundex scores and then finding the differences. Additionally, a limit had to be set in order to decide which words came under a Soundex score. After considerable trial and error, 0.8 was chosen to be the lower limit.

**Algorithm 2: Word Clustering**

**procedure** ClusterWords ()

**Input:** dataset
**Output:** words $w1, w2, …, wN$ are clustered into $c$ number of clusters.

   i. Use the soundex scores that was obtained in Algorithm 1
   ii. Apply DBSCAN clustering algorithm using soundex scores as parameters.
   iii. Experiment with different epsilon values and minimum number of samples to find the optimum number of clusters, i.e, 16.
   iv. Label the clusters by finding majority language (first 2 characters) and category (first 3 characters) in the cluster.
   v. Calculate the clustering coefficient for each cluster.

Similarities between different words are calculated using Jaccard Distance. Different values of Jaccard index were chosen and tested to ultimately select 0.4 as the optimum one. However, these indices were not very relevant to our method and was just another approach that we used. Along with Jaccard, the soundex scores of words, already calculated, was used in clustering the words.

Step 2-4 explains DBSCAN clustering was implemented on the soundex scores of words. With the help of the Trial-and-Error method, we were able to finally decide on these set of parameters - epsilon: 0.0375, min. Samples: 10. With these inputs, we were able to get 16 unique clusters with all similar sounding words put together. The naming convention for the clusters is by using the first 2 characters of the majority language with the first 3 characters of the majority category in the cluster.

The quality of the cluster was measured by calculating a Clustering Coefficient as shown in (1). An undirected graph was created using the words from the dataset, with the vertices being the words and the edges being the levenshtein distance between them. The edges were distance-weighted and a strict threshold of 2 was set.

So, all the edges with a distance value >2 were removed. For calculating the cluster coefficient, the numerator was e/2 (undirected graph) while excluding all the self-loops; the denominator was calculated using the formula = (n*(n-1))/2. The average clustering coefficient was 0.48.

$$\text{coefficient} = e / (n * (n-1)) \quad (1)$$

In (1), e is the number of edges with distance <= 2 and n is the total number of edges.

| Algorithm 3: Language Classification |
|---|
| **procedure** ClassifyWord () |
| **Input:** dataset, DBSCAN cluster |
| **Output:** language |
| i. Divide the dataset into training and testing. |
| ii. Build a custom metric by using Euclidean and Levenshtein distance. |
| iii. Use KNN Classifier to fit your training dataset. |
| iv. Predict the language. |

A custom metric called Linguistic Distance Metric (LDM), shown in (2), is a consolidation of Levenshtein distance for strings and Euclidean distance for numbers (Step ii). The euclidean distance is the most popular distance measure and we used it for finding distance scores between the soundex values. Edit distance (Levenshtein) is used to calculate the distance between 2 strings based on how they are written and this is why the levenshtein distance was used for the words. LDM helps in calculating more refined and accurate distance values. The classifier, using K-Nearest Neighbours, was implemented by taking the Word, Meaning, and corresponding Soundex values as the input (Step iii).

$$d(p, q) = (L(p_{1...i}, q_{1...i}) + ES(p_{1...j}, q_{1...j}))^{1/2} \quad (2)$$

In this equation, p and q are two words, L is the Levenshtein between them and ES is the Euclidean distance between their soundex scores.

Different feature vectors are used to develop the model, and the dataset is used to train it. It gives the Language of a given word as output. Correctly classified words were returned to the clusters in order to strengthen the accuracy of the model. Multiple repetitions helped us achieve an accuracy of 99%.

This research looked at multiclass classification. As a result, using a normal ROC curve would not give the desired outputs. It was, therefore, necessary to use the One vs Rest Classifier to derive the ROC curves for each of the classes.

We summarize below distinctive features of our methodology, challenges we faced, improvisations we did, and any other limitations / caveats related to this work.

i. For this study, the dataset formed the crux. We had to ensure good mix of native words, shared words and loan words for each language in the dataset. For Sanskrit, scraping data from a website was a reasonable option to translate most of the words. There are some words in Sanskrit that do not have equivalent words in other languages. Thus, in case of some words, we have missing values. A snippet of the dataset is given in Table 1.

ii. PyDictionary sometimes presented multiple meanings for a particular word, and manual intervention was necessary to ensure that it associated the correct meaning to the word. Issues relating to polysemy were handled using these techniques.

iii. Clustering coefficient is used as a quality metric, as a proxy for quality of the training sets. This indicates the cohesiveness among words in a given language / word cluster.

iv. Linguistic distance metric - a combination of Levenshtein and Euclidean distance - aiding us to use alphanumeric attributes.

v. Correctly classified words returned to the model in order to strengthen the accuracy in an iterative fashion.

IV. RESULTS

In this section, we present the results that we obtained after analysing the dataset and running the model through the classifier. A snippet of the dataset is shown in Table 1.

TABLE I. DATASET SNIPPET

| English | Hindi | Marathi | Punjabi | Kannada | Tamil | Telugu | Sanskrit |
|---|---|---|---|---|---|---|---|
| above | oopar | varīla | upara | mēle | mēlē | paina | upari |
| appear | dikhaee | disū | pragata | kānisikollu | tōnṟum | kanipiñcē | utplavate |
| act | kaary | kārya | kama | vartisi | nāṭakam | pani | kriyA |
| active | sakriy | sakriya | saragarama | sakriya | ceyalil | kriyāśīla | kriyAzIla |
| activity | gatividhi | kriyākalāpa | saragaramī | catuvatike | natavatikkai | kāryakalāpālu | anuSThA |
| actor | abhineta | abhinētā | abhinētā | nata | natikar | natudu | naTaka |
| art | kala | kalā | kalā | kale | kalai | Kaḷa | kalA |
| again | phir | punhā | dubārā | matte | mīntum | mallī | bhUyas |
| age | aayu | vaya | umara | vayas'su | vayatu | vayas'su | jaraNA |
| answer | uttar | uttara | javāba | uttara | patil | samādhānam | prativAda |

For the clustering module, the jaccard index was used to find words that could be grouped together in terms of similarity of spelling. However, the jaccard index can give different results for various threshold values. In order to reach a final value that would satisfy all clusters without any bias, different indexes were tried out. Table 1 shows a sample word from the clusters for the different indexes.

It is seen that the Jaccard Index 0.2 was very lax and gave an ideal solution. On the other end of the spectrum, 0.5 was too strict and didn't map the words correctly. However, the 0.4 Jaccard Index proved to be the best fit. For the rest of the words, this score was used.

TABLE II. VARIOUS JACCARD INDEXES

| Word | Language | Jaccard Scores | | |
|---|---|---|---|---|
| | | *0.2* | *0.4* | *0.5* |
| peechhe | Hindi | | 0.822 | 0.663 |
| māgē | Marathi | | 1 | 0.723 |
| pichē | Punjabi | 1 | 0.833 | 0.755 |
| hinde | Kannada | | 0.773 | 0.667 |
| pinnāl | Tamil | | 0.721 | 0.648 |
| venuka | Telugu | | 0.833 | 0.617 |
| pRSThataH | Sanskrit | | 1 | 0.777 |

The words were further clustered based on the majority language and category. The categories included prepositions, kinship, people, pronouns, number, anatomy, animals, agriculture, bodily functions, mental functions, nature, directions, fabrication, motion, time, common, adjective, and miscellaneous. For example, man-made materials come under the category of fabrication.

In order to assess the quality of the clusters, clustering coefficients were calculated. Table 2 shows the details for these values.

TABLE III. CLUSTER DETAILS AND COEFFICIENTS

| Cluster Label | Cluster Size | Clustering Coefficient |
|---|---|---|
| Hindi + Fabrication | 1104 | 0.29 |
| Kannada + Fabrication | 310 | 0.7 |
| Kannada + Motion | 196 | 0.7 |
| Kannada + Number | 57 | 0.3 |
| Marathi + Fabrication | 185 | 0.56 |
| Punjabi + Body Functions | 150 | 0.63 |
| Punjabi + Motion | 413 | 0.33 |
| Sanskrit + Fabrication | 51 | 0.42 |
| Sanskrit + Mental Health | 604 | 0.24 |
| Sanskrit + Pronouns | 77 | 0.63 |
| Tamil + Adjectives | 548 | 0.65 |
| Tamil + Fabrication | 1889 | 0.21 |
| Tamil + Motion | 181 | 0.55 |
| Tamil + Pronouns | 255 | 0.28 |
| Telugu + Fabrication | 802 | 0.57 |
| Telugu + Nature | 177 | 0.63 |

The paper presents a multi-step machine learning approach for obtaining language labels, language word groups, and language cluster labels. We have got the root (mother) language of a word belonging to any of these Indian Languages: Sanskrit, Hindi, Punjabi, Marathi, Kannada, Tamil, and Telugu.

In this research, we have proposed a KNN-based solution for supervised classification of Indian Languages.

Using global models like SVMs, ANNs, and decision trees is suitable when there is one decision boundary or when one algorithm can work for all kinds of data instances. For a problem with contrary results, a local model is a validated approach with instance-based learning to be the strongest example. KNN is the most popular algorithm, which is found to give significant results.

The training dataset consists of alpha-numeric features. In such a case, it is hard to rely on one particular metric as input to the KNN classifier. To overcome this obstacle, we introduced a custom metric taking care of both vector distances for numbers and edit distances for strings.

Table 3 shows a snippet of the different results obtained when words were classified into Hindi, Telugu, Punjabi, and Kannada, Tamil, and Sanskrit.

TABLE IV. RESULTS OF WORD CLASSIFICATION

| Word | Meaning | Original | Predicted | Comments |
|---|---|---|---|---|
| phenkana | throwing | Hindi | Telugu | misclassified |
| dhanush | bow | Hindi | Punjabi | common |
| bhavyavada | magnificent | Kannada | Hindi | common |
| āghāta | shock | Kannada | Sanskrit | correct |
| jun'yā | old lady | Marathi | Punjabi | misclassified |
| nirvana | salvation | Marathi | Kannada | common |
| bē'īmān | dishonesty | Punjabi | Hindi | correct |
| pracāra | preaching | Punjabi | Telugu | common |
| nipuṇar | expert | Tamil | Kannada | common |
| aṭakka | compressed | Tamil | Sanskrit | misclassified |
| ākasmika | accidental | Telugu | Sanskrit | correct |
| hatya | murder | Telugu | Kannada | common |
| arogya | health | Sanskrit | Sanskrit | correct |
| mula | root | Sanskrit | Tamil | common |

As is seen, some of the words are common to multiple languages. This was an issue that we faced while building the model and classifying multiple words. In order to get accurate results, it is necessary for us to work on having more data from different languages and a dataset of words that are present in different languages.

Building a ROC curve helped us understand how accurate these results were and if there was scope for better classification in any of these classes. Since this problem is one of multi classification, a One Vs. Rest Classifier was used to create the curves. Table 4 shows the mapping between language and classes along with the values for Area Under the Curve (AUC). Figures 6 and 7 show the results of the ROC curve for these classes.

TABLE V. CLASS AND LANGUAGE MAPPING FOR ROC CURVE

| Class | Language | AUC |
|---|---|---|
| 0 | Tamil | 0.59 |
| 1 | Kannada | 0.52 |
| 2 | Hindi | 0.57 |
| 3 | Marathi | 0.53 |
| 4 | Sanskrit | 0.18 |
| 5 | Punjabi | 0.64 |
| 6 | Telugu | 0.65 |

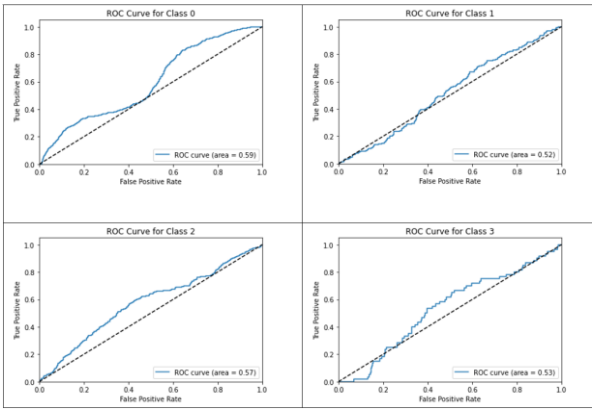

Fig. 4      ROC Curve for Classes 0-3

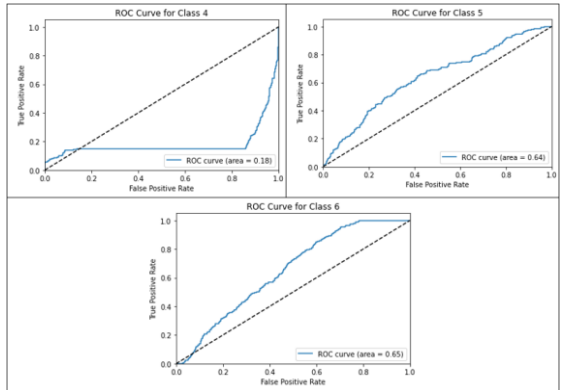

Fig. 5      ROC Curve for Classes 4 - 6

It is seen that the AUC for Class 4 is a very low 0.18. This value needs to be studied more to create concrete conclusions. One plausible reason may be that Sanskrit and each of the Indian languages have lot of common words and thus the data does not follow IID (Independent and Identically Distributed) patterns. However, the other classes have performed relatively well and each of the words can be mapped to the other languages with the assurance that the results are trustworthy. We need to acknowledge that the overlapping of words between languages might affect the classification.

## V. Discussions

The model we have used is a lot more scalable and it grows the data-set by making use of automation (translate APIs). It was also necessary to keep the dataset as free from manual intervention as possible. Crowdsourcing to our associates aided in the corpus completion. Despite that, there were still a few words that could not be translated to Sanskrit; these became part of the missing values. It is self-evolving via iteration between clustering and classification.

Initially, Jaccard index was used for clustering words. But Soundex scores were more relevant given the dataset. Thus, soundex scores were used as input for DBSCAN clustering.

DBSCAN is much preferred because it doesn't need parameter, k, which k-means needs. As opposed to k-means, DBSCAN is robust to noisy points and works well with an imbalanced dataset. As the number of clusters was hidden in the dataset, it was a good choice. It produces a varying number of clusters, based on the input data; this problem lends itself better for application of DBSCAN as there is significant overlap between languages due to shared words. The reachability concept used in DBSCAN makes it more natural to group words into clusters.

We make use of clustering coefficient as a quality metric for our training dataset which is in the form of word clusters. Clustering coefficient should return higher value when a language has retained root word and associated words as is seen in the case of Sanskrit, provided training data is representative. Clustering coefficient can be valuable if we change the granularity of analysis from individual language to groups of languages.

For a given word classification may vary based on the training set we use and the same word may be affected by polysemy and result in mis-classification. In such cases using Ensemble techniques such as Adaboost which synergizes multiple weak classifiers can be particularly useful. Adaboost can be implemented as the learning model, along with KNN. Guo Haixiang et al., have explained that the fundamental design of Adaboost can be used with KNN to classify imbalanced data.

We can also incorporate any rules given by domain experts (from linguistics as well as literature) to build our model. Our model in turn can validate the rules using the quality metric.

## VI. Conclusions

Determining how languages historically evolved will continue to puzzle us. We have made a beginning by breaking away from rigid language family-based approach that presupposes relationships. Our approach enables the classification of words using machine learning. Here, discovery of interrelationship between languages happens automatically. We can further reinforce our approach using the AdaBoost algorithm along with KNN, using boosting-by-resample method.

As our approach enables automated construction of datasets it is not restricted to limited data-sets of cognate words and it can provide full power of data as well as computing. We have also proposed an iterative and evolutionary approach that can incorporate self-learning and self-improvement. We feel our analysis has the potential to take Comparative Linguistics research to an extraordinary scale. An interesting possibility of this exploration is to identify and (possibly) recover lost words in many languages. We intend to submit a research proposal to the Language Computing Group of the Department of Electronics and IT, Government of India to take this work to the next stage.

### Acknowledgment

We would like to show our gratitude to K. S. Srinivas, our professor, PES University for his continued guidance and support.